\pdfoutput=1

\documentclass[11pt]{article}

\usepackage[]{acl}

\usepackage{times}
\usepackage{latexsym}

\usepackage[T1]{fontenc}

\usepackage[utf8]{inputenc}

\usepackage{microtype}

\usepackage{inconsolata}
\usepackage{url}

\usepackage{graphicx}
\usepackage{booktabs}
\usepackage{multirow}
\usepackage{amsmath}
\usepackage{amsfonts}
\usepackage{pifont}
\usepackage{algorithm}
\usepackage{algpseudocode}
\usepackage{hyperref}
\usepackage{tcolorbox} 
\tcbuselibrary{listings} 
\usepackage{makecell}
\usepackage{tabularx}

\title{Success is in the Details: Evaluate and Enhance Details Sensitivity of Code LLMs through Counterfactuals}
\author{Xianzhen Luo$^1$, Qingfu Zhu$^1$\thanks{Corresponding author}, Zhiming Zhang$^1$, Mingzheng Xu$^1$, \\ \textbf{Shijie Xuyang}$^2$, \textbf{Tianhao Cheng}$^2$, \textbf{Yixuan Wang}$^1$, \textbf{Zhiyuan Ma}$^3$, \textbf{Yuantao Fan}$^4$, \textbf{Wanxiang Che}$^1$ \\
$^1$Harbin Institute of Technology, Harbin, China\\
$^2$Fudan University, Shanghai, China \\
$^3$University of Science and Technology of China, Hefei, China \\
$^4$Beijing University of Posts and Telecommunications, Beijing, China \\
\texttt{\{xzluo, qfzhu, zmzhang, car\}@ir.hit.edu.cn}
}

\begin{document}
\maketitle
\begin{abstract}
Code Sensitivity refers to the ability of Code LLMs to recognize and respond to details changes in problem descriptions.
While current code benchmarks and instruction data focus on difficulty and diversity, sensitivity is overlooked. We first introduce the CTF-Code benchmark, constructed using counterfactual perturbations, minimizing input changes while maximizing output changes. 
The evaluation shows that many LLMs have a more than 10\% performance drop compared to the original problems. 
To fully utilize sensitivity, CTF-Instruct, an incremental instruction fine-tuning framework, extends on existing data and uses a selection mechanism to meet the three dimensions of difficulty, diversity, and sensitivity. 
Experiments show that LLMs fine-tuned with CTF-Instruct data achieve over a 2\% improvement on CTF-Code, and more than a 10\% performance boost on LiveCodeBench, validating the feasibility of enhancing LLMs' sensitivity to improve performance.
\end{abstract}

\section{Introduction}

\begin{figure}[t]
\centering
\includegraphics[width=\columnwidth]{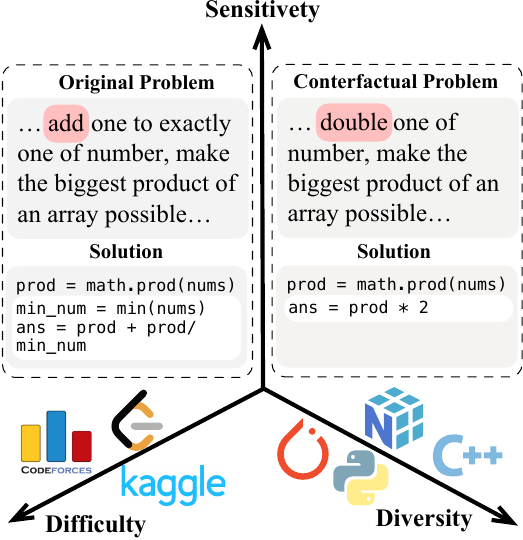} 
\caption{
While diversity and difficulty have been explored, sensitivity to problem details remains underexplored. 
In the original problem, the smallest number should be increased, whereas in the counterfactual version, no matter which number is modified, the result remains the same—double the cumulative sum.
}
\label{fig:intro}
\end{figure}

Code generation is essential for enhancing software engineering efficiency~\cite{zhu-etal-2024-survey}, and also a crucial measure of intelligence~\cite{o1}.
To increase code capabilities, Code Large Language Models (Code LLMs) are developed by pre-training on large-scale code corpora~\cite{qwen25coder, dscoderv2}.
Successful generation requires Code LLMs to accurately map between requirements and algorithmic logic~\cite{10.5555/15670, pressman2005software}. A small mismatch will cause the whole task to fail.
In Figure~\ref{fig:intro}, changing the description from `\textbf{add} one' to `\textbf{double} one' alters the underlying algorithmic logic entirely (see detailed explanation in the figure caption).
As such, the model’s sensitivity to detail becomes a crucial measure of its ability. 

However, the ability of Code LLMs to capture and address such fine-grained differences remains unclear.
Existing code generation benchmarks primarily emphasize difficulty and diversity.
For difficulty, tasks range from basic functions to competition-level algorithms~\cite{evalplus, du2024evaluating, livecodebench}.
For diversity, benchmarks cover data science, system development, and interdisciplinary applications and so many domains~\cite{bigcodebench, lai2023ds, daeval, repobench}.
However, these benchmarks evaluate LLMs on isolated tasks without assessing their sensitivity to subtle differences in requirement details. 
This limitation extends to code instruction fine-tuning. 
Most approaches augment training data along the axes of difficulty and diversity: (1) by incrementally introducing constraints to synthesize harder tasks~\cite{wizardcoder, wang2024dolphcoder},(2) by rewriting or drawing inspiration from real-world code to produce more diverse samples and broaden domain coverage~\cite{magicoder, semi-instruct, wei2024selfcodealign, wavecoder}.
In contrast, constructing datasets that exploit models’ sensitivity remains underexplored.


In response to these gaps, we first introduce the CTF-Code benchmark. 
The inspiration is from counterfactual studies in NLP~\cite{disco, catfood, nl-ctf-survey}, which make minimal changes to inputs to produce outputs that differ substantially.
Concretely, some variations of original problems are sampled at first.
Then, algorithm experts write solutions of solvable variations and select CTF problems which most preserve superficial task similarity and alter the algorithmic logic.
After, the test inputs of original problems are executed on CTF solutions to generate new outputs to construct CTF test cases.
Last, mainstream LLMs are evaluated on the complete CTF-Code benchmark.
Experiments reveal that state-of-the-art models like GPT-4o and Qwen2.5-Coder~\cite{hurst2024gpt, qwen25coder} experience performance drops exceeding 10\% on CTF-Code compared to original problems, highlighting significant `blind spots' in detail sensitivity. 


Furthermore, we introduce the CTF-Instruct pipeline for three-dimensional data construction. 
Starting from an existing dimension (such as difficulty), CTF data are generate to cover the sensitivity dimension. 
Then, a selection mechanism is applied to the sensitivity-enhanced data to complete the third dimension (e.g., diversity). 
Finally, the original base-dimension data are merged with the selected sensitivity data to obtain a dataset that is complete across all three dimensions.
Experiments show that LLMs fine-tuned with CTF-Instruct data achieve a 2.6\% improvement on CTF-Code, and gains on other benchmarks such as HumanEval+ (+4.2\%), BigCodeBench-hard (+5.2\%), and LiveCodeBench (+11.6\%)~\cite{evalplus, bigcodebench, livecodebench}, confirming the help of sensitivity to code abilities.

Our contribution is summarized below:
\begin{itemize}
    \item We propose CTF-Code, the first benchmark focused on sensitivity, and the evaluation results expose the shortcomings of mainstream Code LLMs in understanding requirement details.
    \item  We design a three-dimensional-completed data generation framework, starting from one dimension, completing sensitivity by generation and the last dimension by selection.  
    \item LLMs trained with CTF-Instruct data achieve substantial performance improvements across CTF-Code and other benchmarks compared to existing methods.
\end{itemize}

\section{Related Work}
\paragraph{Code Benchmark} 
Existing code generation benchmarks primarily include two dimensions: (1) Difficulty: from function-level~\cite{mbpp,humaneval}, to class-level~\cite{classeval}, and to contest-level~\cite{livecodebench}; (2) Diversity: BigCodeBench~\cite{bigcodebench} focuses on Python package usage, DS-1000~\cite{ds1000} targets data science, while MultiPLE~\cite{cassano2023multipl} evaluates multilingual code generation. 
However, these benchmarks do not address sensitivity, which evaluates a model’s ability to handle subtle but critical changes in task requirements.
This differs from robustness, which measures the model's ability to produce stable outputs under non-critical changes (e.g., noise or rephrasing) in input~\cite{li2025enhancing,lin2025robunfr,wang2023recode}
In this work, the first sensitivity benchmark, CTF-Code is introduced.

\paragraph{Code Instruction Tuning Datasets}
Most methods on code instruction tuning data augmentation~\cite{self-instruct} mainly focus on difficulty enhancement and diversity expansion. 
\citet{wizardcoder, wizardlm} increase the difficulty of data by adding constraints to seed data~\cite{codealpaca}. 
Considering that the seed data may limit the diversity of generated data, ~\citet{magicoder,wavecoder} rewrite real-world data to better align real distributions, thereby avoiding model bias and enhancing the diversity. 
To combine both dimensions, existing approaches typically adopt multi-stage training~\cite{magicoder,wang2024dolphcoder} or data mixing strategies~\cite{opencode,wavecoder,wu2024inversecoder}.
While these methods have achieved significant success, the usage and combination of sensitivity is overlooked.
\begin{figure*}[t]
\centering
\includegraphics[width=\textwidth]{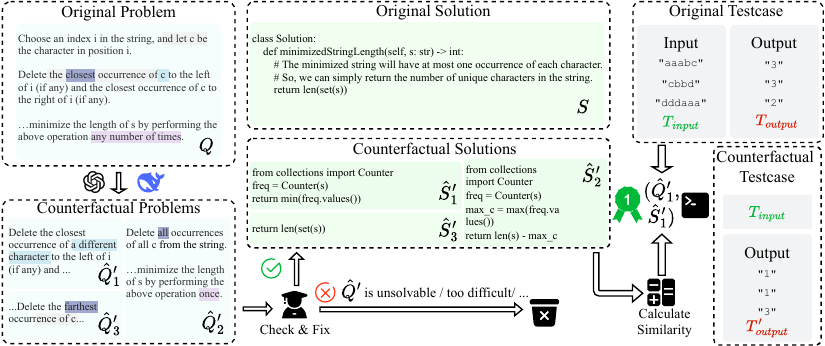} 
\caption{The pipeline of CTF-Code benchmark construction. First, original problems are sent to LLMs to sample semantic permutations on the problem description. Algorithmic experts will carefully check the CTF problems and decide to drop or fix them to generate CTF solutions. After selecting the most suitable CTF problem, its testcases are constructed by executing its solution on the inputs from the original testcases.}
\label{fig:method}
\end{figure*}
\paragraph{Counterfactual in NLP}
Counterfactuals in NLP aim to explore the model's output variation patterns through minimal input perturbations~\cite{realistic-ctf,ctf-ceval, catfood, nl-ctf-survey}. 
Unlike adversarial attacks, which introduce subtle, malicious inputs to mislead the model into producing incorrect or unsafe outputs~\cite{jenko2025blackbox}, counterfactuals refer to make small but critical semantic changes in the input and prompt the model to detect and adjust output accordingly.
Existing studies in the code domain mainly focus on local modifications to code~\cite{hoodalarge}, such as flipping conditional statements or modifying logical operators, testing the model's ability to differentiate and understand counterfactual code~\cite{gu2024counterfeit,cito2022counterfactual}. 
To the best of our knowledge, no prior work has leveraged counterfactual perturbations at the problem level to study models' sensitivity to requirement details.

\section{CTF-Code Benchmark}
\subsection{Formal Definition}
Evaluation for code tasks is test-driven, with its basic unit formalized as a tuple $\mathcal{P} = (Q, T, S)$,  where $Q$ is the problem description,  $T = \{t_i\}_{i=1}^n$ is a set of test cases, with each $t_i = (\text{input}_i, \text{output}_i)$, and $S$ is a solution that satisfies 
$\forall t_i \in T, S(\text{input}_i) = \text{output}_i$.
Based on this, the goal of generating counterfactual problem can be formulated as an optimization problem. Given the original problem $\mathcal{P}$, generate $\mathcal{P}' = (Q', T', S')$ such that
\begin{equation}
\begin{aligned}
& \underset{Q',S'}{\text{maximize}} & & \mathcal{D}_S(S, S') \\
& \text{subject to} & & \mathcal{D}_Q(Q, Q') \leq \epsilon
\end{aligned}
\label{eq:1}
\end{equation}
where $\mathcal{D}_Q$ is the description similarity function, $\mathcal{D}_S$ is the solution difference function, and $\epsilon$ is the similarity threshold.
We use the normalized Levenshtein distance~\cite{levenshtein1966binary} as $\mathcal{D}_Q$, and define $\mathcal{D}_S$ as one minus the cosine similarity between code embeddings.
This optimization objective ensures that $Q'$ is highly similar to $Q$, while $S'$ and $S$ differ significantly. After obtaining $Q'$ and $S'$, the new test cases $T'$  are constructed.

\subsection{Benchmark Construction}
As shown in Figure~\ref{fig:method}, the construction of CTF-Code follows a three-phase paradigm: First, select problems that have a large semantic space as the original problem $\mathcal{P}$. Then, apply semantic perturbations to generate CTF description $Q'$ and derive the CTF pairs $Q', S'$ based on the optimization objective. Finally, construct the new test cases $T'$ while ensuring no data bias.

\begin{table}[t]
\centering
\begin{tabular}{lccc}
\toprule
 & \textbf{Acc.} & \textbf{Problems} & \textbf{Length} \\
\midrule
Humaneval & \textbf{96.3} & 164 & 71.6  \\
LCB-Easy & 95.6 & \textbf{215} & \textbf{210.5} \\
\bottomrule
\end{tabular}
\caption{Comparison between HumanEval and LiveCodeBench (LCB) -Easy. \textbf{Acc.} represents the Pass@1 score of o1-mini on both benchmarks. \textbf{Problems} indicates the number of problems, \textbf{Length} represents the average word count per problem description.}
\label{tab:compare_he_lcb}
\end{table}

\paragraph{Original Data Selection} The easy subset of LiveCodeBench (LCB) ~\cite{livecodebench} is selected as $\mathcal{P}$. 
Table~\ref{tab:compare_he_lcb} shows that LLMs can solve nearly all problems in this subset, minimizing the impact of difficulty. 
Additionally, the problem length is 210.5 words, could give more semantic space for perturbations. 
Furthermore, algorithmic competition problems require participants to carefully consider every detail and boundary condition, where even small deviations can lead to wrong answers. 
This aligns perfectly with the goal of sensitivity. 

\paragraph{CTF Pair Generation} This step aims to generate \( Q' \) and \( S' \). 
Given the complexity, a heuristic generation-selection strategy is proposed to approximate the solution of Equation~\ref{eq:1}. 
Based on \(Q\), several LLMs sample \(K\) candidates \(\{\hat{Q}'_k\}_{k=1}^K\) using the prompt in Figure~\ref{fig:prompt_1} in Appendix~\ref{app:prompt}.
Specifically, after comparing to other LLMs, the best-performing LLMs, including \texttt{gpt-4o}, \texttt{gpt-4turbo}, and \texttt{o1-mini}, each generate five samples, as further sampling primarily yielded duplicates. 
After reviewing all $\hat{Q}'_k$, we empirically set $\epsilon=0.13$, which balances formal similarity with allowance for semantic divergence.
Only $\hat{Q}'_k$ satisfied $\mathcal{D}_Q(Q, \hat{Q}'_k) \leq \epsilon$ are retained.

However, retained $\hat{Q}'_k$ may be unsolvable or too difficult.
Four competition programmers are invited to annotate, each of whom has at least a bronze medal in ICPC~\footnote{\href{https://icpc.global/}{International Collegiate Programming Contest}}.
The detailed annotation process is in Appendix~\ref{sec:anno}.
Annotators are required to read \(\hat{Q}'_k\) and judge:(1)solvability, (2) if it is a CTF problem (filter problems which different descriptions yielding identical solutions), and (3) if its difficulty changed from \( Q' \). 
Prior to the annotation, a 10-problem trial is conducted to ensure annotator consistency.
Each problem is then independently annotated by two programmers. 
Where annotations disagree, a third annotator provides a new judgment, and the outcome is determined by majority vote.
For the passed \(\hat{Q}'_k\), annotators then write \(\hat{S}'_k\).
The pair $(\hat{Q}'_k, \hat{S}'_k)$ that maximizes Equation~\ref{eq:2} is selected as $(Q', S')$.
\begin{equation}
\underset{(\hat{Q}'_k, \hat{S}'_k)}{\arg\max} \left[ \mathcal{D}_S(S, \hat{S}'_k) - \lambda \mathcal{D}_Q(Q, \hat{Q}'_k) \right].
\label{eq:2}
\end{equation}
\(\lambda\) is a scaling factor that ensures \(\mathcal{D}_S\) and \(\mathcal{D}_Q\) can compute. 
It is set as $1.2$. 
Through this heuristic rule, we obtain an approximate optimal \( Q' \) and \( S' \).

\paragraph{CTF Testcase Completion} To ensure the performance change of LLMs latter only from details change between \( (Q, Q') \), a dual-constraint test case generation mechanism is designed to avoid the influence from \( (T, T') \). \textbf{Input Space Inheritance}: We retain the original testcases’ input distribution, i.e., \( T'_{\text{input}} = T_{\text{input}} = \{input_i\}_{i=1}^n \).
\textbf{Output Space Reconstruction}: The expected output is generated based on the new solution \( S' \), i.e., for each \( \text{input}_i \in T_{\text{input}} \), \( \text{output}'_i = S'(\text{input}_i) \). Finally, \( T' = \{(\text{input}_i, \text{output}'_i)\}_{i=1}^n \) is constructed. 
The data distribution interference is eliminated by fixing the input variables,
and the correctness of the test case is ensured by the correctness of \(S'\).  
Additionally, fixed inputs enable backtracking when the LLM behavior differs between \(Q, Q'\).

Compared to the traditional code benchmarks that evaluate isolated problems, CTF-Code introduces paired data with only details differences to enable analysis of sensitivity for the first time, as shown in Figure~\ref{fig:original problem} and Figure~\ref{fig:ctf problem 1}. 
Ultimately, CTF-Code curated a set of 186 problems.

\section{CTF-Instruct}
Unlike difficulty and diversity, detail sensitivity has not been explored in existing instruction datasets.
To address the gap, an incremental data construction approach, CTF-Instruct, is proposed.
Starting with datasets that satisfy a single dimension (e.g., difficulty), sensitivity data are generated through counterfactual perturbations. 
Then, a selection algorithm based on the third dimension (diversity) is applied, ultimately constructing a dataset that cover all three dimensions.
\begin{figure*}[t]
\centering
\includegraphics[width=\textwidth]{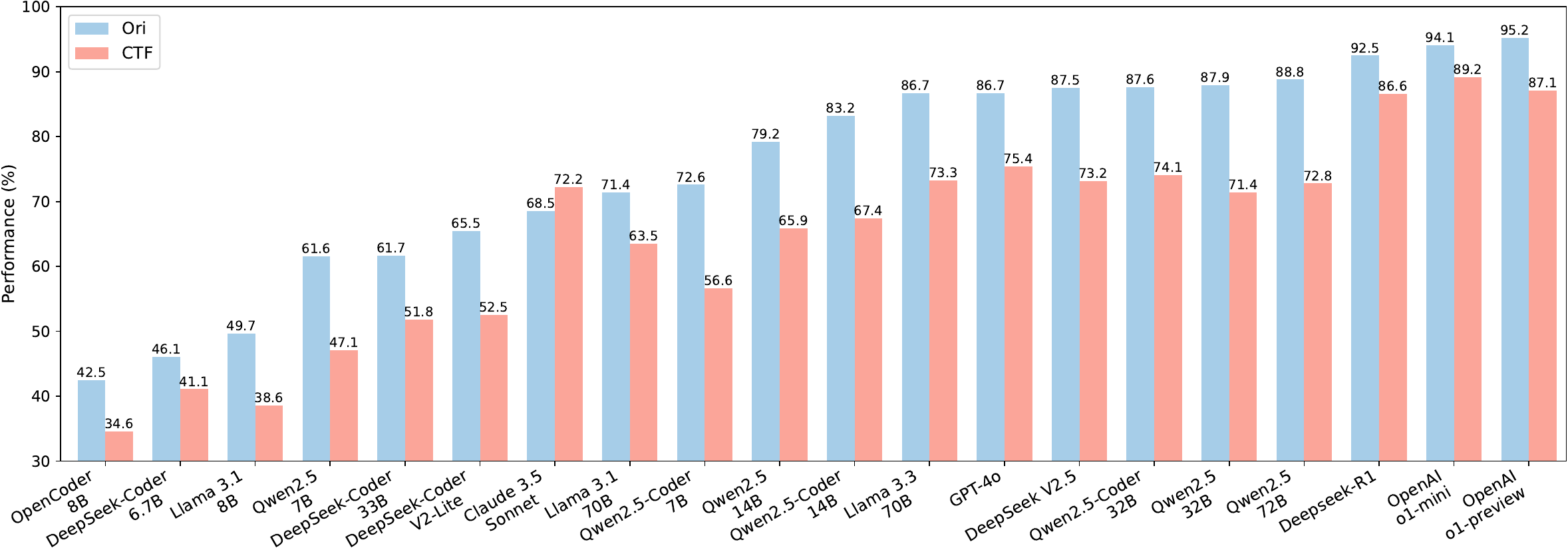} 
\caption{The evaluation results of Code LLMs on CTF-Code.}
\label{fig: CTF-Code Benchmark}
\end{figure*}
\subsection{Generation}
\label{sec:generation}

We first tried generating paired sensitivity data from scratch, but we found that the generated problems are too easy and repetitive. 
As the existence of high-quality data like Evol-Instruct (110k), which satisfies the difficulty dimension, 
incrementally expanding the data is more effective and efficient. 
Prompt~\ref{fig:prompt_2} with \texttt{gpt-4-turbo} is applied to generate sensitivity pair \(\mathcal{D}_{sens}\) based on the difficulty data \(\mathcal{D}_{diff}\).
After generation, duplicates found in existing benchmarks are removed to avoid data leakage following~\citet{wizardcoder}. 

\begin{table}[ht]
\centering
\begin{tabular}{lcc}
\toprule
Percentile & Similarity & Difficulty Difference \\
\midrule
25\% & 0.99 & 0.00 \\
50\% & 0.97 & 0.01 \\
75\% & 0.92 & 0.14 \\
95\% & 0.64 & 0.89 \\
\bottomrule
\end{tabular}
\caption{Percentile values for the semantic embedding similarity and difficulty difference distributions of original and sensitive data. The Similarity column indicates the percentage of values greater than the given threshold, while the Difficulty Difference column represents the percentage of values less than the given threshold.}
\label{tab:similarity_difficulty_distributions}
\end{table}


The 102k generated \(\mathcal{D}_{sens}\) are evaluated on difficulty and diversity.
For difficulty, we follow~\citet{wang2024your} and use their trained scorer to assign 1–5 scores to $Q\in\mathcal{D}_{diff}$, $Q'\in\mathcal{D}_{sens}$. 
99\% of the absolute score difference of $Q'$ and its seed $Q$ is less than 1 in Table~\ref{tab:similarity_difficulty_distributions}, indicating minimal difficulty shift.
For diversity, we compute the cosine similarity between embeddings of $(Q, S)$ and $(Q', S')$, extracted by DeepSeek-Coder 1.3B~\cite{guo2024deepseek}. 
The model is tuned on Code-Feedback~\cite{opencode} for one epoch for alignment. Each embedding is the average of the final-layer token hidden states of $S$.
As shown in Table~\ref{tab:similarity_difficulty_distributions}, 75\% of samples have similarity above 0.92.
These results suggest that counterfactual generation preserves difficulty and diversity, and more importantly, that the sensitivity dimension is relatively independent of the other two dimensions.

\subsection{Selection}
\label{sec:selection}

Since \(\mathcal{D}_{sens}\) is generated based on \(\mathcal{D}_{diff}\), they naturally share similar diversity limitations. Directly merging them would amplify this bias.
However, we can select a subset of \(\mathcal{D}_{sens}\) that maximizes diversity relative to \(\mathcal{D}_{diff}\). 
This introduces a distribution shift, partially mitigating the diversity deficiency.
We use the semantic embeddings computed in Section~\ref{sec:generation}, and apply the k-center greedy (Algorithm~\ref{Alg:kcentergreedy}) to select the most diverse subset of \(\mathcal{D}_{sens}\).
Notably, some outlier samples (meaningless or corrupted) exhibit extremely large semantic distances. These are excluded by sorting based on distance and removing the tail.
We empirically set the subset size $|\mathcal{D}_{sens}^{sub}|=30$k.
By merging $\mathcal{D}_{sens}^{sub}$ with $\mathcal{D}_{diff}$, we obtain the dataset CTF-Instruct that satisfies all three dimensions. 
An example of this distributional adjustment is shown in Figure~\ref{fig:plot} in Appendix~\ref{sec:app_exp}.

A similar process is applied when starting from diversity data, Oss-Instruct (75k).
After 73k sensitivity data generated by \texttt{gpt-3.5-turbo}, we apply the difficulty scorer in Section~\ref{sec:generation} and retain only the 10k-size subset with the highest difficulty scores.
After combined the subset with the Oss-Instruct, we get CTF-Instruct\textsubscript{oss}. 
In both cases, sensitive data is first generated from the existing dimension, and a selection algorithm is used to fill the remaining missing dimension.

\begin{table*}[t]
\centering
\begin{tabular}{cc|c|cc|cc|cc}
\toprule
 \multirow{2}{*}{Base} & \multirow{2}{*}{Model} & \multicolumn{1}{c|}{EvalPlus} & \multicolumn{2}{c|}{LiveCodeBench} 
 & \multicolumn{2}{c|}{BigCodeBench} & \multicolumn{2}{c}{CTF-Code}\\
\cmidrule{3-9}
 & & HumanEval (+) & All & Easy & Full & Hard & Ori & CTF\\
\cmidrule{1-9}
\multirow{6}{*}{\makecell{DeepSeek\\Coder 6.7B}}
& DC-6.7B-Instruct & 74.4 (71.3) & 18.9 & 45.3 & 35.5 & 10.1 & 45.8 & 38.1\\

& Wavecoder & 75.0 (69.5) & 18.9 & 46.0 & 33.9 & 12.8 & 47.7 & 39.2\\

& Inversecoder & 76.2 (72.0)  & 18.1 & 43.1 & \multicolumn{1}{r}{35.9} & 10.8 & 47.8 & 39.1\\

& Magicoder & 76.8 (71.3) & 19.2 & 46.6 & 36.2 & 13.5 & 48.8 & 43.4\\

& CTFCoder & \textbf{78.7} (\textbf{75.0}) & \textbf{21.4} & \textbf{53.3} & \textbf{37.6} & \textbf{14.2} & \textbf{52.8} & \textbf{44.5}\\
& CTFCoder\textsubscript{oss} & 71.3 (65.9) & 18.3 & 46.6 & 37.0 & 12.2 & 51.4 & 43.1 \\
\cmidrule{1-9}
\multirow{6}{*}{\makecell{Qwen2.5\\Coder 14B}}
& Evol & 85.4 (79.3) & 23.9 & 71.5 & 43.7 & 14.2 & 76.3 & 59.9 \\
& CTF & \textbf{88.4} (\textbf{80.5}) & \textbf{24.6} & \textbf{74.1} & 44.1
& \textbf{17.6} & \textbf{79.5} & \textbf{60.8} \\
& w/o select & 85.4 (78.0) & 24.1 & 72.8 & \textbf{44.2} & 16.2 & 76.4 & 60.2 \\
\cmidrule{2-9}
& Oss & 84.1 (77.4) & 20.6 & 61.8 & 42.0 & 12.2 & 75.5 & 58.6 \\
& CTF\textsubscript{oss} & 86.0 (79.9)  & \textbf{22.3} & \textbf{67.9} & \textbf{42.5} & \textbf{18.9} 
& \textbf{78.2} & \textbf{60.0} \\
& w/o select & \textbf{86.6} (\textbf{80.5}) & 20.8 & 67.2 & \textbf{42.5} & 14.9 & 76.8 & 59.2\\
\bottomrule
\end{tabular}
\caption{
Performance comparison of CTFCoder with other models. To avoid environmental discrepancies, the official leaderboard results are presented. Only when results are missing, local testing are conducted. 
`w/o select' means original data mix random selected sensitive data, without methods in Section~\ref{sec:selection}.
}
\label{tab:main_results}
\end{table*}

\section{Experiment}
\subsection{CTF-Code Benchmark}

\paragraph{Models} We evaluate Qwen 2.5 Coder, Deepseek Coder v1 \& 2, OpenCoder, Qwen 2.5, Llama 3.1 \& 3.3, GPT-4o (\texttt{gpt-4o-2024-08-06}), Claude 3.5 Sonnet (\texttt{claude-3.5-sonnet-20240620}), o1-mini, o1-preview and Deepseek-R1~\cite{qwen25coder,guo2024deepseek,qwen2025qwen25,zhu2024deepseek,huang2024opencoder,grattafiori2024llama,guo2025deepseek}.


\paragraph{Evaluation}

As shown in Figure \ref{fig: CTF-Code Benchmark}, most LLMs exhibit a significant performance drop on CTF-Code, often exceeding 15\%.
Reasoning-oriented LLMs such as R1 and O1 experience notably smaller drops, suggesting a stronger ability to capture fine-grained variations in problem requirements. 
This gap is especially obvious in problems that can be simplified or transformed. 
Reasoning LLMs tend to abstract key properties to reformulate the problem, whereas other LLMs are more likely to mimic the problem description, often becoming misled by the counterfactual phrasing.
For LLMs families such as Qwen2.5-Coder, we observe that the sensitivity gap narrows with increasing model size, indicating a positive correlation between model scale and the sensitive ability.
Interestingly, Claude-3.5-Sonnet even outperforms its original performance on CTF-Code, highlighting its strong generalization capabilities and practical robustness in code-related scenarios.

To further understand these results, we analyze common failure cases. 
The most frequent error is that models fail to recognize the semantic change in the CTF variant and instead solve it as if it were the original problem. 
This may be due to that the original or similar problems exist in the LLM’s training data. 
Even when LLMs capture the details change, their performance often degrades on solvable yet uncommon CTF variants. 
Common issues include incorrect ordering of logical operators in \texttt{if} statements, confusion between data structures (e.g., lists and sets), and failure to handle boundary conditions.
These problems are especially frequent in tasks that require case-by-case reasoning or involve numerous conditional branches.

Overall, our findings suggest that current LLMs still have substantial room for improvement in sensitivity to details. 
Misinterpreting such details not only leads to incorrect solutions but also disrupts the generation process itself. 
Enhancing sensitivity remains a crucial direction for advancing the performance and reliability of code LLMs.

\subsection{Instruction Tuning}
\paragraph{Setup}
CTFCoder and CTFCoder\textsubscript{oss} are obtained from using CTF-Instruct, CTF-Instruct\textsubscript{oss}, respectively, finetuned on Deepseek Coder 6.7B base for 3 epochs. Qwen 2.5 Coder 14B Base~\cite{qwen25coder} is also tuned.
During training, the batch size is 512 and the sequence length is 2048. The initial learning rate is \textnormal{2e-5} with 10 warmup steps, and the learning rate scheduler is cosine.

\begin{table*}[h]
\centering
\begin{tabular}{lcccccccc}
\toprule
Model & C\# & C++ & Java & PHP & TypeScript & Bash & JavaScript & Avg \\
\midrule
DC-6.7b-Instruct & 67.7 & 66.5 & \textbf{69.0} & 46.6 & 70.4 & 41.8 & 73.9 & 62.3 \\
Wavecoder        & 69.0 & 57.8 & \textbf{69.0} & 52.2 & \textbf{74.2} & 39.9 & 70.8 & 61.8 \\
Inversecoder     & 69.6 & 68.3 & 63.9 & 41.6 & 72.3 & 43.3 & 73.3 & 61.8 \\
Magicoder        & 67.7 & \textbf{69.6} & 65.8 & 44.7 & 69.2 & 41.1 & 72.0 & 61.4 \\
CTFCoder         & \textbf{72.2} & 67.1 & 65.2 & \textbf{53.4} & 73.0 & \textbf{43.7} & \textbf{74.5} & \textbf{64.2} \\
\bottomrule
\end{tabular}
\caption{Performance comparison of various LLMs on different programming languages in MultiPLE.}
\label{tab:multiple}
\end{table*}

\paragraph{Baseline \& Benchmark}
Other LLMs tuned on Deepseek Coder 6.7B Base are compared, including Deepseek Coder 6.7B Instruct~\cite{guo2024deepseek}, Magicoder~\cite{magicoder}, Wavecoder~\cite{wavecoder}, and Inversecoder~\cite{inversecoder}. 
Qwen 2.5 Coder finetuned on the original Evol-Instruct and Oss-Instruct are the baselines.

The benchmarks cover a range of difficulty levels, including Humaneval(+)~\cite{humaneval, evalplus}, and LiveCodeBench~\cite{livecodebench}. Humaneval+ adds a lot of test cases to Humaneval to cover corner cases. LiveCodeBench collects algorithm problems from Online Judges and 
includes three difficulty levels: easy, medium, and hard. Since GPT-4-turbo’s training data ends in December 2023, we test LiveCodeBench questions after January 2024. For diversity, BigCodeBench~\cite{bigcodebench} is selected for Python package usage, and MultiPLE is for multilingual generation, including C\#, C++, Java, PHP, TypeScript, Bash, and JavaScript. Additionally, BigCodeBench selects high-difficulty sub-data to form a Hard subset.

\subsection{Results}
Table~\ref{tab:main_results} shows the performance comparison between CTFCoder and other LLMs. CTFCoder demonstrates consistent performance improvements across all benchmarks. Although previous models already cover difficulty and diversity and achieve strong performance, the addition of sensitivity acts like a further ``activation''. 
CTFCoder shows significant improvements across all three dimensions. 
On sensitivity, it has a nearly 3\% improvement on CTF-Code, indicating that CTF indeed helps the model pay more attention to details. On difficulty, Humaneval+, BigCodeBench-Hard, and LiveCodeBench, CTFCoder achieves over 4\%, 5\%, and 11\% performance improvements, respectively. 
On diversity, although CTF-Instruct is not explicitly designed for multilingual programming, it exhibits strong cross-language generalization.
CTFCoder achieves the best performance on C\#, PHP, Bash, and JavaScript in MultiPL-E, with a notable improvement of nearly 4\% on C\# and an average gain of 3\% across languages in Table~\ref{tab:multiple}.
Combined with results on BigCodeBench, these demonstrate that CTFCoder generalizes well across diverse domains and programming languages.

CTF-Instruct, building upon the difficulty dimension of Evol-Instruct, results in comprehensive enhancement. This illustrates that generating sensitivity data using existing data as seeds not only preserves the original data dimensions but can even trigger further improvements. 

Even though CTFCoder\textsubscript{oss} has a relatively small amount of SFT data, CTF\textsubscript{oss} helps it outperform other models on LiveCodeBench-Easy and BigCodeBench-Full, reflecting the `activation' effect on diversity works, too. 
On Qwen 2.5 Coder 14B, compare the baseline, random selection of CTF-Instruct data (`w/o select') and CTF generally shows a progressive performance improvement, highlighting the effectiveness of sensitivity data and the importance of data selection.

\section{Discussion}

\begin{figure}[t]
\centering
\includegraphics[width=0.9\columnwidth]{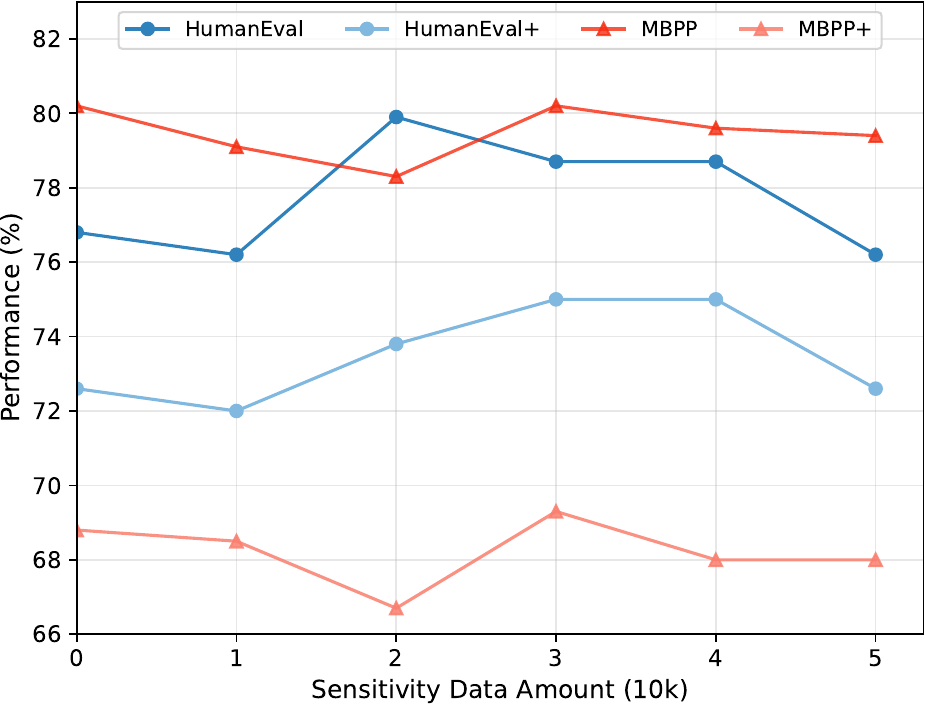} 
\caption{
The change in model performance as sensitivity data joined into Evol-Instruct.
}
\label{fig:DC-Evol}
\end{figure}

\textbf{There exists an optimal range for the amount of sensitivity data.} Figure~\ref{fig:DC-Evol} shows the performance trend when sensitivity data is gradually mixed into Evol-Instruct (110K), with the performance evolving in three stages: an initial decline, a mid-stage increase, and a final decline. The initial drop indicates that a certain amount of sensitive data is required to have an effect. The subsequent rise followed by a decline suggests that there is an upper limit for sensitivity data, confirming our observation that directly merging sensitivity and original data dimensions exacerbates the lack of the third dimension. Figure~\ref{fig:DC-OSS} in Appendix~\ref{sec:app_exp} also shows the results for Oss-Instruct (75k). However, it does not exhibit an initial performance drop. This may be because the diversity-oriented data is relatively easy to learn, and thus the addition of sensitive data does not introduce huge interference. However, when too much sensitive data is added, a decline similar to that observed before emerges.

\begin{figure}[t]
\centering
\includegraphics[width=0.9\columnwidth]{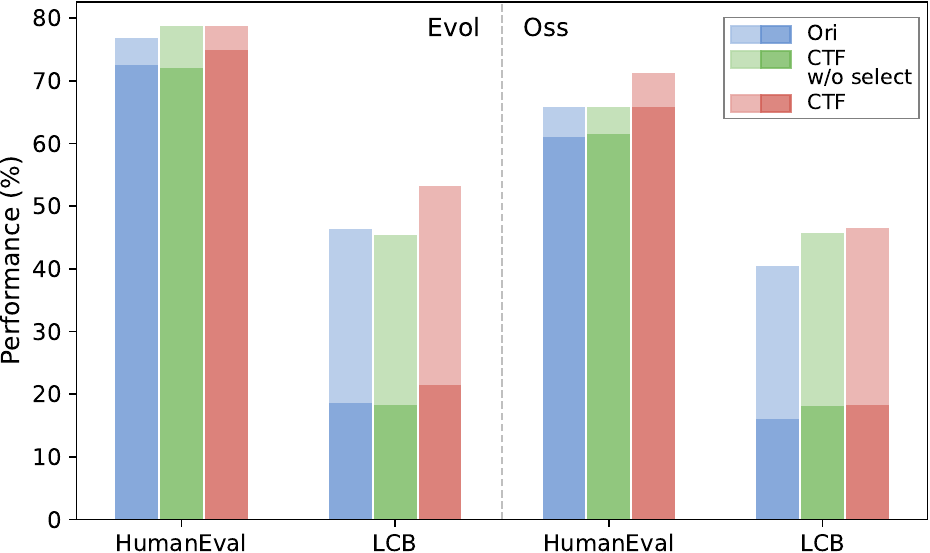} 
\caption{
The performance change brought by the selection strategy on Evol-Instruct and Oss-Instruct. The darker shade in Humaneval represents Humaneval+, while in LCB, the darker shade represents LCB-All, and the lighter shade represents LCB-Easy.
}
\label{fig:selection}
\end{figure}


\textbf{The effectiveness of the selection strategy is universal.} Table~\ref{tab:main_results} and Figure~\ref{fig:selection} compare the performance of different models and data using the selection strategy versus not using it (`w/o select') with the same amount of data. Regardless of the original data or base model, the strategy generally leads to performance improvement. Figure~\ref{fig:selection} shows that, with Evol-Instruct, performance on LiveCodeBench improved by over 17\%, while for OSS-Instruct, performance on Humaneval increased by more than 7\% compared to `w/o select'. This validates our hypothesis that data offset can effectively address the third dimension.



\begin{table}[t]
\centering
\begin{tabular}{cccc}
\toprule
Epoch & Strategy & HE (+) & LCB (Easy) \\
\cmidrule{1-4}
\multirow{4}{*}{2} & 2+0 & 74.4 (69.5) & 18.5 (46.8) \\
& 1+1 
& \textbf{79.9} (\textbf{75.0}) & \textbf{21.0} (\textbf{51.8}) \\
\cmidrule{2-4}
& 2+0 & 65.9 (61.0) & 16.0 (40.4)\\
& 1+1 & \textbf{66.5} (\textbf{61.6})  & \textbf{18.8} (\textbf{46.9}) \\
\cmidrule{1-4}

\multirow{4}{*}{3} & 3+0 & \textbf{76.8} (72.6) & 18.6 (46.4) \\
& 2+1 
& \textbf{76.8} (\textbf{73.2}) & \textbf{20.7} (\textbf{51.3}) \\
\cmidrule{2-4}
& 3+0 & 65.2 (59.8) & 17.0 (42.5) \\
& 2+1 & \textbf{68.3} (\textbf{62.2})  & \textbf{19.0} (\textbf{47.2})\\
\bottomrule
\end{tabular}
\caption{
The results of continual training with CTF. Epoch is the total number of training epochs, and `x+y’ indicates that the model is first trained for \(x\) epochs on the original data, followed by \(y\) epochs on CTF-Instruct. HE represents Humaneval, and LCB refers to LiveCodeBench-All, with `Easy' inside the parentheses.
}
\label{tab:continual_train}
\end{table}

\begin{figure}[t]
\centering
\includegraphics[width=0.9\columnwidth]{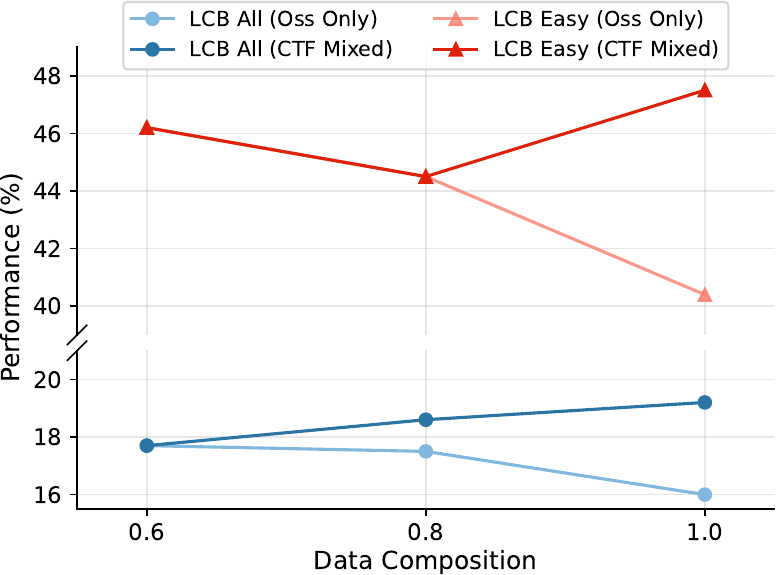} 
\caption{
Performance changes with the increase in diversity data (Oss only) and the gradual injection of sensitivity into the diversity data (CTF Mixed).
}
\label{fig:data_amount}
\end{figure}

\textbf{Under a fixed data amount, incorporating sensitive data still brings improvements.}
Since the amount of data used for training the open-source models in the main experiment differs, we designed a controlled experiment to verify the independent gain from the sensitivity dimension. In Figure~\ref{fig:data_amount}, when the total training data volume is fixed, replacing 40\% of the original Oss-instruct data with randomly selected CTF data led to a 20\% improvement on LCB-All, whereas simply increasing the Oss-instruct data volume caused a 9.5\% performance drop. 

\textbf{Sensitivity can be directly used for continual training.} Inspired by Magicoder~\cite{magicoder}, in Table~\ref{tab:continual_train}, after training on the original data for 1 or 2 epochs, an additional epoch of CTF-Instruct is added. Compared to continuing training with the original data alone, this approach shows a significant performance improvement. Particularly on LiveCodeBench, every setup achieves a 10\% gain. This further demonstrates the orthogonality of the sensitivity dimension with the other two dimensions, as its benefit does not depend on joint training, allowing for efficient and convenient continual training to achieve gains.
\section{Conclusion}
Beyond diversity and difficulty, we introduced sensitivity as a key dimension for evaluating and improving Code LLMs. By constructing the CTF-Code benchmark, we revealed the shortcomings of existing Code LLMs in understanding details. To futher utilize sensitivity, we propose the CTF-Instruct framework, which generates sensitivity data based on existing dimensions to cover sensitivity and employs a filtering algorithm to shift towards the third dimension. Experiments show that CTF-Instruct data fine-tuned LLMs improves performance on CTF-Code and outperform existing open-source models on general code generation benchmarks, validating the universal benefits of sensitivity optimization for Code LLMs.

\clearpage
\section*{Limitation}
Due to constraints in training resources and manpower, our work was limited to constructing a relatively modest set of CTF-Code problems, without exploring the potential for more complex or challenging examples. Additionally, the CTF-Instruct framework was not tested with multi-round generation, nor was it evaluated on larger, more advanced LLMs. While our experiments demonstrate the effectiveness of the proposed approach on the models tested, we acknowledge that the full potential of CTF-Instruct could be realized by scaling up the dataset and conducting more extensive fine-tuning experiments, particularly on models with greater capacity. Furthermore, the impact of training on larger models with more rounds of fine-tuning remains an open question and is a promising direction for future work.
\section*{Ethical Considerations}
The data for the proposed methods is drawn solely from publicly accessible project resources on reputable websites, ensuring that no sensitive information is included. Moreover, all datasets and baseline models used in our experiments are also available to the public. We have taken care to acknowledge the original authors by properly citing their work.



\bibliography{custom}
\newtcblisting{promptbox}{
    colback=white, 
    colframe=black, 
    boxrule=0.5pt, 
    arc=0pt, 
    left=6pt, 
    right=6pt, 
    top=4pt, 
    bottom=4pt, 
    boxsep=4pt, 
    width=\textwidth, 
    listing only, 
}

\clearpage
\appendix
\section*{Appendix}
\label{sec:appendix}
\renewcommand{\thesection}{\Alph{section}}
\section{Supplement For CTF-Instruct}
\label{sec:app_exp}

\begin{algorithm}[]
\caption{K-Center Greedy Selection}
\begin{algorithmic}[1]
\State \textbf{Input:} Sensitivity data \( \mathcal{D}_{sens} \), needed data amount \( \tau \), original data \( \mathcal{D}_{base} \)
\State \textbf{Output:} Set \( \mathcal{D}_{sub} \subseteq \mathcal{D}_{sens} \) of \( \tau \) data
\State \( C \gets \mathcal{D}_{base} \) \Comment{Initialize centers}
\For{$i = 1$ to $k$}
    \State \( \text{dist}_x \gets \min_{y \in \mathcal{D}_{base} \cup \mathcal{D}_{sub}} ||\phi(x) - \phi(y)||_2 \)
    \State \( x \gets \arg\max_{x \in \mathcal{D}_{sens}} \, \text{dist}_x \)
    \Comment{Select the farthest data \( x \)}
    \State \(\mathcal{D}_{sub} \gets \mathcal{D}_{sub} \cup \{x\} \)\Comment{Update centers}
    \State \(\mathcal{D}_{sens} \gets \mathcal{D}_{sens} - \{x\} \) \Comment{Update data}
\EndFor
\State \textbf{Return} \( \mathcal{D}_{sub} \) \Comment{Return the set of \( \tau \) data}
\end{algorithmic}
\label{Alg:kcentergreedy}
\end{algorithm}

\begin{figure}[]
\centering
\includegraphics[width=\columnwidth]{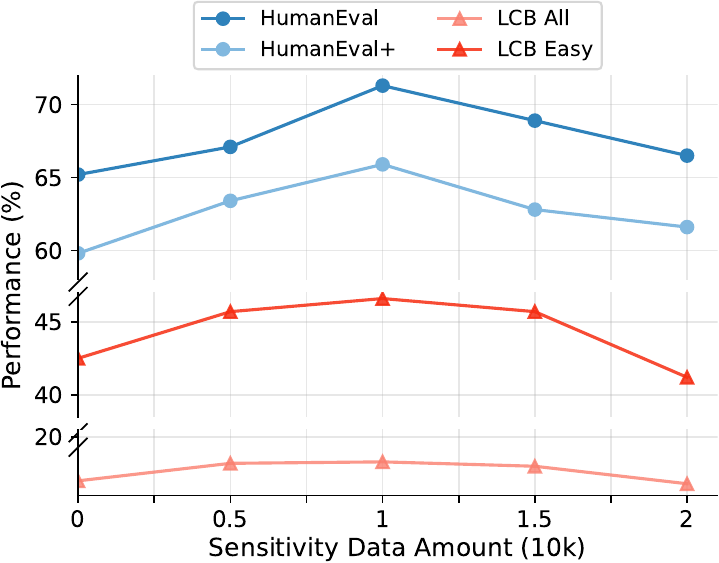} 
\caption{
The performance varies with the amount of sensitivity data mixed into OSS-Instruct.
}
\label{fig:DC-OSS}
\end{figure}
\begin{figure*}[t]
\centering
\includegraphics[width=\textwidth]{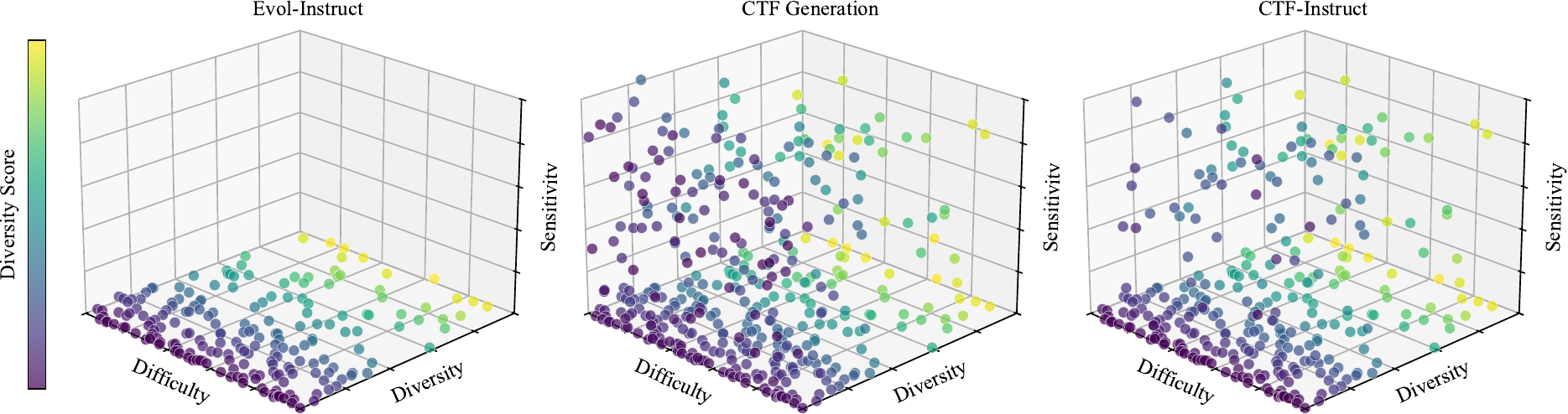} 
\caption{
The data distribution change trace during the CTF-Instruct pipeline.
}
\label{fig:plot}
\end{figure*}

\begin{figure*}[]
\centering
\includegraphics[width=\textwidth]{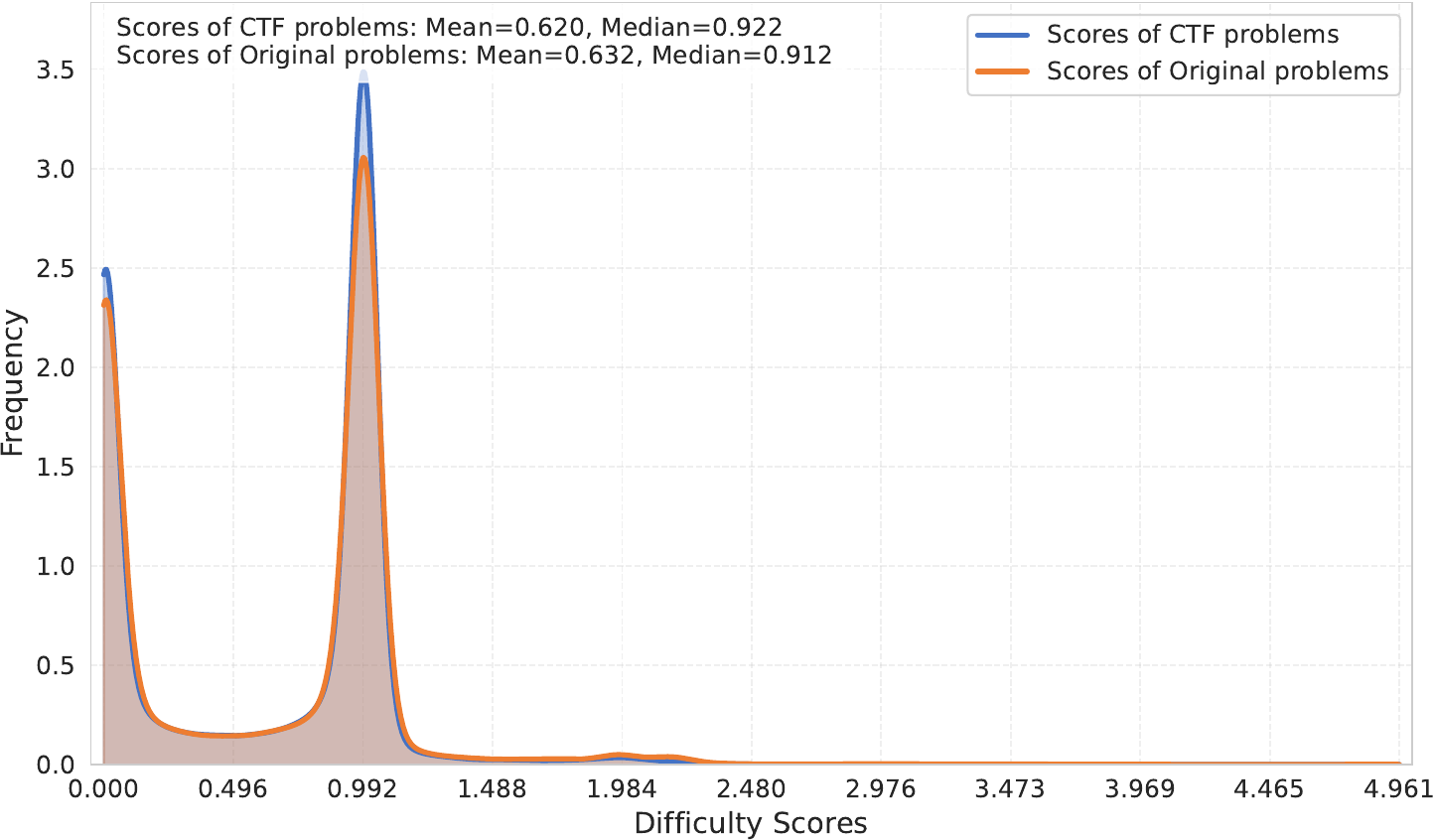} 
\caption{
The distribution of difficulty scores of sensitive data and its original data.
}
\label{fig:diff_scores}
\end{figure*}
\begin{figure*}[]
\centering
\includegraphics[width=\textwidth]{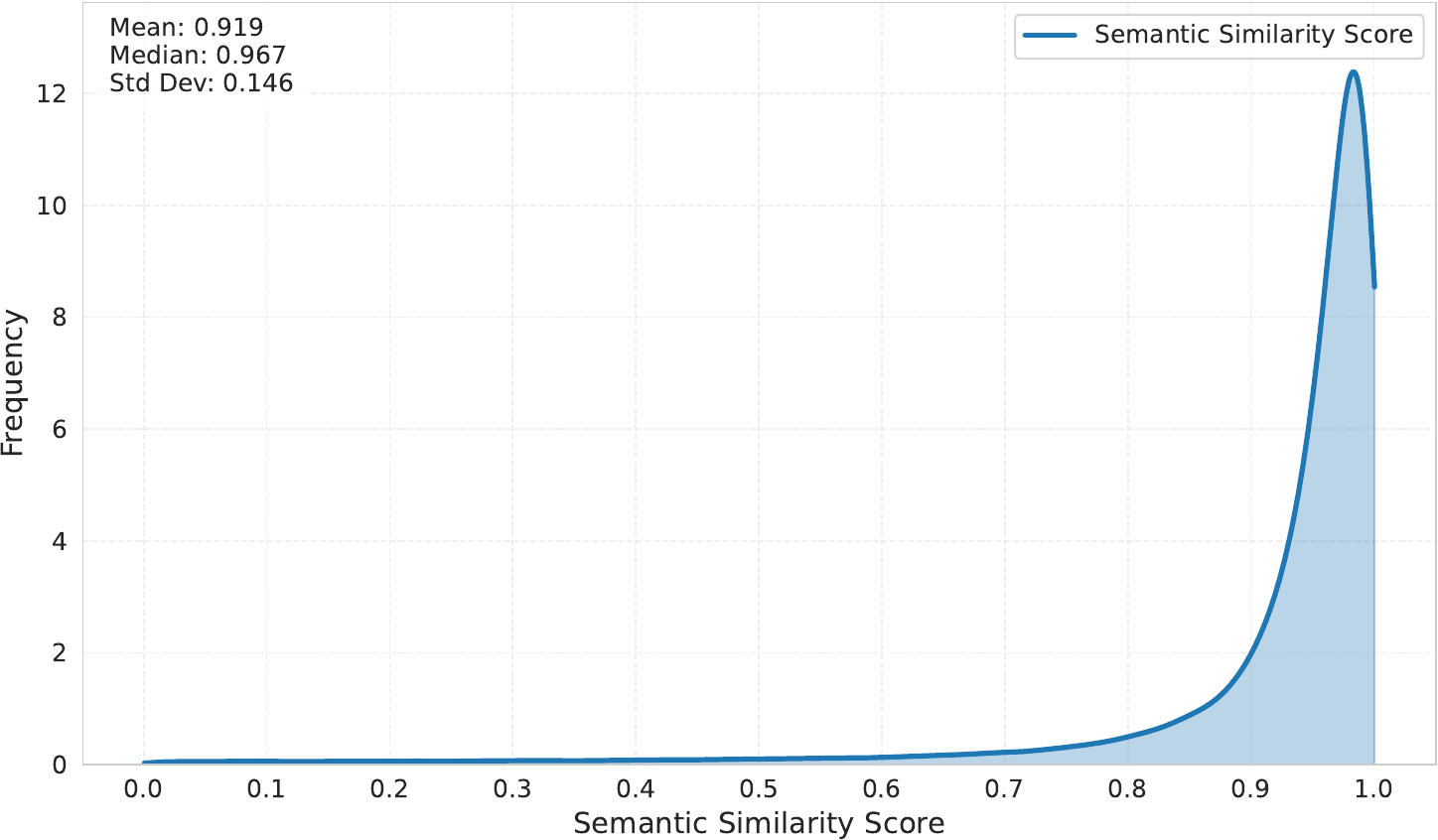} 
\caption{
The distribution of the semantic embedding similarity scores of sensitive data and its original data.
}
\label{fig:seman_simi}
\end{figure*}

\clearpage
\section{Annotation Process}
\label{sec:anno}
\subsection{Mission Background}
Today, large language models (LLMs) demonstrate remarkable capabilities in code generation. However, it remains unclear how well LLMs can capture the nuances of programming problem details, such as the distinction between ``swapping any two characters'' and ``swapping two adjacent characters''. Can LLMs accurately capture the differences between these two concepts? 
To investigate this, we propose to modify a set of original problems (LeetCode Easy Level) to construct a new set of \textbf{counterfactual} (\textbf{CTF}) problems. 
These CTF problems are designed to have minor textual differences from the original problems while yielding significantly different solutions. 
To avoid the bias from test cases, we aim to ensure that CTF problems can utilize the test cases of the original problems without the need for reconstruction.

\subsection{Annotation Content}
The annotation process does not involve modifying the problem itself, as this task has already been done by the LLMs. 
Instead, the annotator's role is simply to evaluate whether the modified problem is correct and aligns with our requirements.

\subsection{Construction Workflow}
To illustrate the construction workflow in detail, we will supplement it with an example.
\begin{enumerate}
  \item Read the original problem and briefly explain the meaning of the original problem. As shown in Figure~\ref{fig:original problem}, the meaning of the original problem is: "Given a string consisting of three letters \texttt{'abc'} in any order, can \texttt{'abc'} appear after swapping any two characters at most once?"
  \item Read and understand the newly automatically generated problem. If there are errors in the \texttt{Sample Input/Output} or in the \texttt{Test Cases}, correct them.
  \item In comparison with the original problem, classify the new problem into three types (Bad, Robust, CTF) and explain what changes have been made.
  \begin{itemize}
    \item Bad. The new problem has a significant vulnerability (logical vulnerability or conflict) and can not be a complete problem.
    \item Robust. The new problem has only a different wording from the original question, i.e. the algorithm used by the new problem and the answer is exactly the same. As shown in Figure~\ref{fig:robust problem}, this is a robust version of the original problem. After understand the meaning of the new problem, we can tell that the change is "any substring can be reversed".\\
    For the new problem, the total length of cards is 3. Reversing a substring of length 3 is equivalent to swapping the letters in positions \texttt{1} and \texttt{3}, and position \texttt{2} will not be changed during the reversing process; reversing a substring of length 2 is equivalent to swapping adjacent letters in the original question. The operation of the original problem and the operation of the new problem are exactly the same. Therefore, the answers are completely consistent and do not need to be modified.
    \item CTF. The new problem has only a small difference from the original problem, but it changes the meaning of the original problem, making the answers not exactly the same as the original problem (With not too much variation in difficulty, the more variation in answers the better).\\
    Figure~\ref{fig:ctf problem 1} and Figure~\ref{fig:ctf problem 2} are two example of CTF problems. The change of the former problem is "only two adjacent characters can be exchanged", and the change of the latter problem is "cards become \texttt{abcd}".
  \end{itemize}
  \item Determine whether new test cases need to be added to the CTF problem. For example, the annotator should determine whether the range of data of the new problem is fully consistent with the original problem, and whether the input of test cases of the original problem can be directly executed by the CTF problem.\\
  For the first CTF problem, there is no need to add new test cases, while for the second CTF problem, some new test cases should be added.
\end{enumerate}

\subsection{Annotation Tabular}
As shown in Table~\ref{tab:annotation}, we provide an example of the annotation table that the annotator should fill in.

\begin{table*}[ht]
\centering
\small
\begin{tabularx}{\textwidth}{c|X|c|c|c|c|X|c}
\toprule
\makecell{Original\\Problem\\Index} 
& \makecell{Original Problem\\Meaning} & Model 
& \makecell{New\\Problem\\Index} 
& \makecell{New Problem\\Statement\\Error} 
& \makecell{New\\Problem\\Type} 
& \makecell{Modification} & \makecell{Add New\\Test Cases}\\
\cmidrule{1-8}
0 & Given a string composed of letters 'abc' in any order, exchange any two characters to see if string 'abc' can occur.
& o1-mini & 0-0 & & Robust 
&  & No \\

\cmidrule{1-8}
0 & Given a string composed of letters 'abc' in any order, exchange any two characters to see if string 'abc' can occur.
& o1-mini & 0-1 & & CTF 
& Only two adjacent characters can be exchanged & No \\

\cmidrule{1-8}
1 & Add 1 to a number in an array of positive numbers, how to maximise the array product
& o1-mini & 1-1 & & CTF 
& Replace a number in an array with a number from 0-9, how to make the array product maximum & No \\

\cmidrule{1-8}
4 & A string with a phone number in front and 2 digits in the middle indicating age. Find those over 60 years old
& o1-mini & 4-0 & & CTF 
& How many people are over 60 years old and have unique phone numbers? & Yes \\

\cmidrule{1-8}
4 & A string with a phone number in front and 2 digits in the middle indicating age. Find those over 60 years old
& o1-mini & 4-1 & & CTF 
& Age is hexadecimal & No \\

\bottomrule
\end{tabularx}
\caption{An example of the annotation table.}
\label{tab:annotation}
\end{table*}

\begin{figure*}[htbp]
    \centering
    \begin{promptbox}
## Question Content:

There are three cards with letters $\texttt{a}$, $\texttt{b}$, $\texttt{c}$ placed in a row in some order. You can do the following operation at most once: 

 
-  Pick two cards, and swap them.  Is it possible that the row becomes $\texttt{abc}$ after the operation? Output "YES" if it is possible, and "NO" otherwise.

Input

The first line contains a single integer $t$ ($1 \leq t \leq 6$) - the number of test cases.

The only line of each test case contains a single string consisting of each of the three characters $\texttt{a}$, $\texttt{b}$, and $\texttt{c}$ exactly once, representing the cards.

Output

For each test case, output "YES" if you can make the row $\texttt{abc}$ with at most one operation, or "NO" otherwise.

You can output the answer in any case (for example, the strings "yEs", "yes", "Yes" and "YES" will be recognized as a positive answer).Sample Input 1:
6

abc

acb

bac

bca

cab

cba

Sample Output 1:

YES
YES
YES
NO
NO
YES

Note

In the first test case, we don't need to do any operations, since the row is already $\texttt{abc}$.

In the second test case, we can swap $\texttt{c}$ and $\texttt{b}$: $\texttt{acb} \to \texttt{abc}$.

In the third test case, we can swap $\texttt{b}$ and $\texttt{a}$: $\texttt{bac} \to \texttt{abc}$.

In the fourth test case, it is impossible to make $\texttt{abc}$ using at most one operation.

## Starter Code:

## Test Cases:

"[{\"input\": \"6\\nabc\\nacb\\nbac\\nbca\\ncab\\ncba\\n\", \"output\": \"YES\\nYES\\nYES\\nNO\\nNO\\nYES\\n\", \"testtype\": \"stdin\"}]"

\end{promptbox}
    \caption{An example of the original problem.} 
    \label{fig:original problem} 
\end{figure*}

\begin{figure*}[htbp]
    \centering
    \begin{promptbox}
## Question Content:

There are three cards with letters $\texttt{a}$, $\texttt{b}$, $\texttt{c}$ placed in a row in some order. You can perform the following operation at most once:

- Choose any substring of the cards and reverse it.

Is it possible that the row becomes $\texttt{abc}$ after the operation? Output "YES" if it is possible, and "NO" otherwise.
...

\end{promptbox}
    \caption{An example of the robust version of the original problem.} 
    \label{fig:robust problem} 
\end{figure*}

\begin{figure*}[htbp]
    \centering
    \begin{promptbox}
## Question Content:
There are three cards with letters $\texttt{a}$, $\texttt{b}$, $\texttt{c}$ placed in a row in some order. You can perform the following operation at most once:

- Pick two **adjacent** cards and swap them.

Is it possible that the row becomes $\texttt{abc}$ after the operation? Output "YES" if it is possible, and "NO" otherwise.
...
\end{promptbox}
    \caption{The first example of the CTF version of the original problem.} 
    \label{fig:ctf problem 1} 
\end{figure*}

\begin{figure*}[htbp]
    \centering
    \begin{promptbox}
## Question Content:

There are four cards with letters $\texttt{a}$, $\texttt{b}$, $\texttt{c}$, $\texttt{d}$ placed in a row in some order. You can do the following operation at most once:

- Pick two cards, and swap them.  Is it possible that the row becomes $\texttt{abcd}$ after the operation? Output "YES" if it is possible, and "NO" otherwise.
...
\end{promptbox}
    \caption{The second example of the CTF version of the original problem.} 
    \label{fig:ctf problem 2} 
\end{figure*}

\clearpage
\section{Prompt}
\label{app:prompt}
This section shows the prompt used to instruct LLMs to generate desired counterfactual question and instruction tuning data.

\begin{figure*}[htbp]
    \centering
    \begin{promptbox}
Please create a **counterfactual** version of the given original python programming problem. Your goal is to **make a minimal change to the problem that leads to a significant change in the solution**. Follow these detailed steps:
1. Carefully read and comprehend the original problem's context, conditions, constraints, and requirements.
2. Identify a critical point in the original problem and think about a modification. **The modification should be slight but cause a substantial change in the solution approach**.
3. Consider the influence of the modification. Ask yourself: Would it change data structures or algorithms? Explain the influence before output the counterfactual problem. If the influence does not impact the solution approach significantly, rethink another critical point to modify. Repeat Step 2 and Step 3 until you find a point that satisfies the requirement.
4. Modify the original problem based on the most influential point. The modified problem must be consistent, clear, and requires a significantly different solution approach. Update the sample inputs and outputs to match the new problem condition.
5. Output the counterfactual problem, ensuring the following format:
    - Before the JSON format, include a section marker "###Counterfactual Problem".
    - After the section marker, provide the counterfactual problem in the same JSON format as the original, including "question_content", "starter_code", "public_test_cases", and "metadata".
### Original Problem
\end{promptbox}
    \caption{The prompt used to generate CTF-Code Problem.} 
    \label{fig:prompt_1} 
\end{figure*}

\begin{figure*}[htbp]
    \centering
    \begin{promptbox}
Refine a code generation task, initially presented as #Original_Sample#, which is a JSON dict including three keys: a task instruction, and the output generated from  the instruction.
Your task is to produce a #Modified_Sample# by altering the original task instruction in a way that significantly changes the output, yet with minimal adjustments to the instruction itself.

## Requirements:
1. **Minimal Instruction Change**: Achieve the code change with minimal alterations to the instruction. The difference will be assessed through evaluated by the Rouge score, indicating the high similarity in wording, sentence structure, and length to the original.
2. **No Trival Changes to Instruction**: Ensure the modification to the instruction is semantic-relevant. Do not make trivial changes like adding or removing a word, changing the order of words, or replacing synonyms.
3. **Maximal Code Change**: Your adjustments should lead to considerable changes in the output, impacting aspects like algorithms, data structures, data and control flows, or boundary conditions. The difference will be assessed through both the Rouge score and AST score, indicating the output's functionality, implementation, and naming should substantially diverge from the original.
4. **Encourage Trival Code Change**: The code output should be significantly different. Change every aspect of the code, including the function name, variable names.

## Format:
1. Your output should be a #Modified_Sample# dict in **JSON format** as the #Original_Sample# is.
2. Using **markdown code snippet syntax** in the instruction and the output.
3. Ensure all characters are **properly escaped** in the JSON string.

## Examples:
{seeds}

## Question:
- Original_Sample:
\end{promptbox}
    \caption{The prompt used to generate CTF-Instruct data.} 
    \label{fig:prompt_2} 
\end{figure*}

\clearpage



\end{document}